\documentclass[10pt,twocolumn,letterpaper]{article}

\usepackage[pagenumbers]{cvpr}      
\definecolor{cvprblue}{rgb}{0.21,0.49,0.74}
\usepackage[pagebackref,breaklinks,colorlinks,allcolors=cvprblue]{hyperref}
\usepackage{multirow}
\usepackage{algorithm}
\usepackage{algorithmic}
\newcommand{\resultstablefont}{\footnotesize}
\newcommand{\resultstableformat}{%
  \resultstablefont
  \setlength{\tabcolsep}{3pt}%
  \renewcommand{\arraystretch}{1.08}%
}
\newcommand{\mronehead}{\shortstack{MR\\@1 $\uparrow$}}
\newcommand{\rmseonehead}{\shortstack{RMSE\\@1 $\downarrow$}}
\newcommand{\mrtwentyhead}{\shortstack{MR\\@20 $\uparrow$}}
\newcommand{\rmsetwentyhead}{\shortstack{RMSE\\@20 $\downarrow$}}
\def\paperID{*****} 
\def\confName{CVPR}
\def\confYear{2026}

\title{CrystalGRPO: Target-Aligned and Coverage-Preserving Reinforcement Learning for Flow-Based Crystal Structure Prediction}

\author{
\parbox{\textwidth}{
\centering
Kaixiang Su\textsuperscript{1}
\qquad
Hongfei Xue\textsuperscript{1,*}
\qquad
Qiang Zhu\textsuperscript{2,3,*}
\\[5pt]
\textsuperscript{1}Department of Computer Science,
University of North Carolina at Charlotte
\\
\textsuperscript{2}Department of Mechanical Engineering and Engineering Science,
University of North Carolina at Charlotte
\\
\textsuperscript{3}North Carolina Battery Complexity, Autonomous Vehicle and
Electrification (BATT CAVE) Research Center
\\
Charlotte, NC 28223, USA
\\[4pt]
{\ttfamily\small
ksu4@charlotte.edu
\quad
hongfei.xue@charlotte.edu
\quad
qzhu8@charlotte.edu
}
\\[3pt]
\textsuperscript{*}Corresponding authors
}
}

\begin{document}
\maketitle
\begin{abstract}
Flow-based generative models can efficiently produce candidate structures for crystal structure prediction (CSP), but their pretrained objectives do not directly optimize downstream target recovery. Reinforcement-learning post-training offers a flexible solution, yet existing approaches rely primarily on energy rewards and coordinate-only stochastic policies. Predicted energy does not identify the reference polymorph, while reward-driven concentration can reduce the candidate coverage required for Top-$N$ recovery. We introduce \textbf{CrystalGRPO}, a CSP-aligned post-training framework that extends existing ODE-to-SDE policy constructions to the joint coordinate--lattice state. CrystalGRPO combines MACE-predicted energy with a \texttt{StructureMatcher}-based recovery score and provides two operating modes: \textbf{CrystalGRPO-Q}, which prioritizes single-draw recovery, and \textbf{CrystalGRPO-C}, which combines full-trajectory reference regularization with a coverage-aware group advantage to preserve finite-budget target recovery. 
Across MP-20 and MPTS-52 with PXRDGen and OMatG backbones, both variants reduce one- and twenty-sample RMSE relative to coordinate-only reinforcement in all four backbone--dataset settings. CrystalGRPO-Q consistently improves Top-1, whereas CrystalGRPO-C achieves a higher Top-20 across all settings. 
\end{abstract}    
\section{Introduction}
\label{sec:intro}

Crystal structure prediction~\cite{metni2026generative} (CSP) aims to recover the lattice and atomic arrangement of a crystal from its chemical composition and experimental observations~\cite{li2025powder}. In practice, a generative CSP model serves as a proposal engine: it should produce a compact set of candidates that can subsequently be verified by higher-fidelity calculations or experimental refinement~\cite{nikhar2022reliable}. We therefore view CSP as budgeted target recovery, where both single-candidate accuracy and target coverage under a finite candidate budget matter.

\begin{figure}[t]
\centering
\includegraphics[
    width=\columnwidth,
    trim=0 0 0 0,
    clip
]{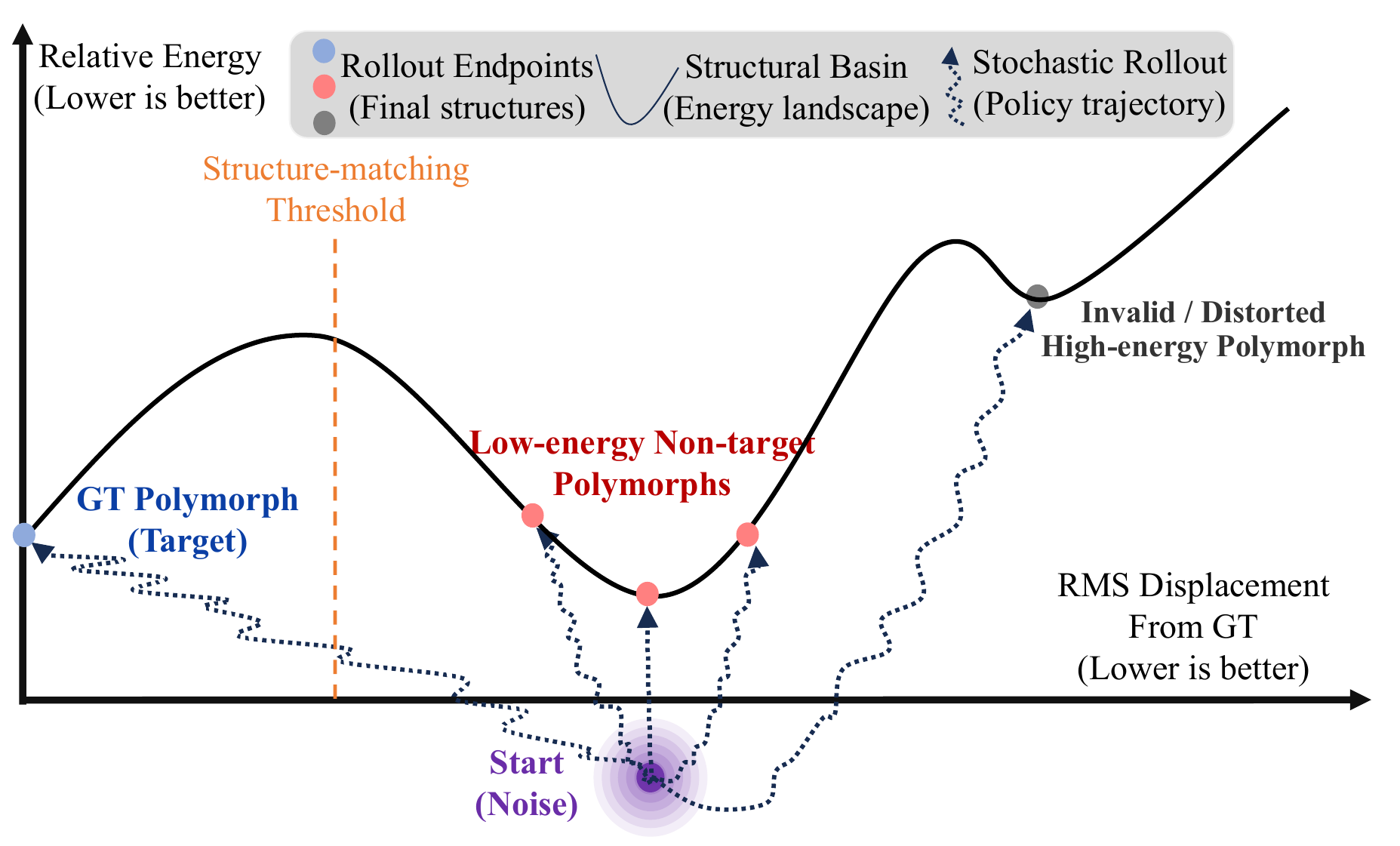}
\caption{\textbf{Predicted energy does not uniquely identify the target polymorph in crystal structure prediction.} This schematic projection places stochastic rollout endpoints according to their predicted relative energy and structural displacement from the reference structure. Rollouts starting from latent noise may terminate in the reference-matched basin (blue), valid low-energy non-target basins (red), or an invalid or distorted high-energy regions (gray). Because a non-target polymorph can be assigned lower predicted energy than the reference, energy-only reinforcement may favor non-target endpoints despite their structural mismatch.
}
\label{fig1}
\end{figure}

Generative models are well suited to this setting because they learn multimodal distributions over periodic structures and can efficiently produce multiple candidates for the same composition. Diffusion~\cite{jiao2023crystal} and flow-matching models~\cite{lipman2022flow,miller2024flowmm} have consequently achieved strong performance on standard CSP benchmarks, with flow models offering particularly efficient sampling~\cite{luo2025crystalflow}. Their pretraining objectives, however, primarily reproduce the distribution of known structures rather than directly optimizing the downstream criteria used to select candidates. Reinforcement learning~\cite{schulman2017proximal} (RL) offers a modular post-training interface: it reuses the pretrained structural prior while directly optimizing non-differentiable terminal rewards without differentiating through the evaluator. Recent work~\cite{liu2025flowgrpo} has made such optimization technically feasible for flow models by equipping deterministic ODE samplers with stochastic transitions and tractable policy likelihoods. These advances establish how flow models can be exposed to policy optimization, but they do not resolve what the policy should optimize for candidate-based CSP.

A natural reward is predicted energy. Energy provides a dense, physics-informed ranking signal that penalizes unfavorable structures and favors candidates with better predicted stability, making it particularly suitable for de novo materials generation \cite{chen2025matinvent}. However, composition-conditioned CSP poses a different objective: recovering the particular polymorph represented by the target observation. Multiple physically plausible polymorphs of the same composition can have similar predicted energies \cite{sun2016thermodynamic, martirossyan2026all}, and an approximate interatomic potential need not assign the lowest energy to the observed structure. As illustrated in Fig.~\ref{fig1}, energy-based reinforcement may therefore favor a low-energy non-target basin over the reference polymorph. Energy provides useful physical guidance, but not target identity. A CSP-aligned reward must combine it with an explicit target-recovery signal.

The candidate-set view of CSP exposes a second mismatch. Standard group-relative optimization repeatedly shifts probability mass toward candidates with higher relative rewards, which can concentrate the rollout distribution around a small number of modes and suppress other physically plausible basins \cite{uehara2024fine}. This concentration may improve single-draw recovery (i.e., top-1 accuracy) while reducing the probability that a set of 20 candidates contains the target structure. Reward-guided CSP must therefore improve target alignment without sacrificing the coverage required for Top-N recovery.

To address these challenges, we propose CrystalGRPO, a CSP-aligned reinforcement-learning framework for post-training flow-based crystal generators. First, building on existing stochastic policy constructions for flow models, CrystalGRPO extends stochasticization and reward-based credit assignment to the joint coordinate–lattice state. Fractional-coordinate transitions respect the periodic torus, lattice transitions remain Euclidean.
This joint treatment contrasts with coordinate-only reinforcement \cite{omatg-rl}, which integrates the lattice deterministically with the frozen pretrained process.
Second, CrystalGRPO uses a hybrid reward in which predicted energy provides dense physics-informed guidance, while a StructureMatcher-based recovery score identifies the reference polymorph and rewards geometrically accurate matches. Third, CrystalGRPO introduces a two-level coverage-preservation mechanism: full trajectory reference regularization limits policy drift during coarse-basin selection, and a coverage-aware advantage prevents valid but unmatched candidates from being suppressed solely because they fall below the group mean. These components define two operating modes of the same framework. CrystalGRPO-Q prioritizes single-draw target recovery, whereas CrystalGRPO-C activates coverage preservation to favor recovery from a finite candidate set.

We evaluate CrystalGRPO on MP-20 and MPTS-52 with two pretrained backbones. As a pre-RL control, we first empirically verify that replacing deterministic ODE sampling with the corresponding stochastic sampler leaves MR@1 and MR@20 nearly unchanged, suggesting that the ODE-to-SDE conversion largely preserves the target-recovery behavior of the pretrained flow. The improvements observed after post-training can therefore be attributed to reward-guided policy updates rather than to the change of sampler itself. Compared with coordinate-only reinforcement, joint coordinate--lattice optimization achieves lower RMSE at both the single-sample and 20-sample levels across all four backbone--dataset settings. It also shows more consistent recovery behavior across backbones: CrystalGRPO-Q achieves across-the-board improvements in MR@1, while CrystalGRPO-C achieves a higher MR@20 across all settings. 
Our analysis further shows that, under composition-only conditioning, energy-based reinforcement can increase both target matches and low-energy non-target candidates. This ambiguity is substantially reduced when PXRD conditioning is provided, indicating that target-specific information makes energy a more reliable refinement signal. Together, these results suggest that finite-budget CSP benefits from joint structural optimization, target-aware guidance, and explicit preservation of the candidate coverage learned during pretraining.

Our contributions are threefold:
(1) We formulate reward-guided CSP around two distinct requirements—target alignment and finite-budget candidate coverage. To address both, we introduce a hybrid reward and coverage-preserving objective that provides complementary single-draw and coverage-oriented operating modes.
(2) Building on existing flow-policy constructions, we extend stochastic transitions and policy optimization to the joint coordinate–lattice state, allowing both fields to receive stochastic exploration and policy-gradient updates.
(3) Across two datasets and two pretrained backbones, we demonstrate that joint coordinate--lattice state provides more consistent gains than coordinate-only reinforcement, while CrystalGRPO-Q and CrystalGRPO-C offer complementary improvements in single-sample and finite-budget recovery. PXRD-conditioned analysis further reveals how target-specific information reshapes the interaction among energy guidance, target identification, and candidate coverage.

\section{Related Work}
\paragraph{Generative models for crystalline materials.}
Early deep generative approaches such as CDVAE
\cite{xie2022crystal} modeled periodic crystal structures with
variational autoencoders, while DiffCSP \cite{jiao2023crystal}
introduced joint diffusion over fractional coordinates and lattices.
Subsequent diffusion models incorporated stronger crystallographic
inductive biases: DiffCSP++ \cite{jiao2024space} imposes explicit
space-group constraints, while SymmCD \cite{levy2025symmcd},
WyckoffDiff \cite{kelvinius2025wyckoffdiff}, and LEGO-xtal
\cite{ridwan2026ai} incorporate symmetry or target local-environment
information into the generation process. Beyond standard CSP,
MatterGen \cite{MatterGen2025} scales diffusion-based crystal
generation and supports property-conditioned inverse design.
In parallel, autoregressive language models represent crystals as text
or CIF sequences, enabling complete-structure generation and the
learning of structural priors
\cite{gruver2024fine,antunes2024crystal,khastagir2026llm}.
Alternative continuous-time formulations have also emerged.
CrysBFN \cite{wu2025periodic} introduces a periodic Bayesian flow for
crystal generation, whereas FlowMM \cite{miller2024flowmm} and
CrystalFlow \cite{luo2025crystalflow} use flow matching to model
periodic coordinates and lattice variables with efficient sampling.
OMatG \cite{omatg} adopts stochastic interpolants, a broader framework
that includes diffusion and flow matching as special cases.
PXRDGen \cite{li2025powder} builds on FlowMM and introduces PXRD
conditioning, supporting both composition-only and
diffraction-conditioned CSP within a unified architecture. We use
PXRDGen and OMatG as pretrained backbones to evaluate whether RL
post-training transfers across distinct continuous-time generative
formulations.

\paragraph{Reinforcement learning for crystal generation.} Existing work has focused mainly on de novo generation (DNG), where composition and structure are generated jointly. MatInvent \cite{chen2025matinvent} and Chemeleon2 \cite{park2026guiding} apply policy-gradient methods to diffusion and latent diffusion models respectively to optimize stability objectives. PackFlow \cite{subramanian2026packflow} addresses organic molecular crystals and bypasses exact policy probabilities through reward weighting and flow loss surrogates. OMatG-IRL \cite{omatg-rl} first apply policy-based RL to CSP, which perturbs the ODE directly to construct a surrogate stochastic process for coordinate-only reinforcement. In contrast, CrystalGRPO jointly reinforces coordinates and lattices and explicitly preserves candidate coverage under finite-budget target recovery.
\section{Preliminaries}
\label{sec:preliminaries}

In this section, we introduce key concepts underlying flow-based crystal structure prediction \cite{miller2024flowmm,li2025powder} and
GRPO for flow-matching models \cite{liu2025flowgrpo}.

\paragraph{Flow-based crystal structure prediction.}A periodic crystal is represented as
$\mathcal{M}=(A,F,L)$, where $A=(a_1,\ldots,a_N)$ denotes
the atomic species, $F\in\mathbb{T}^{N\times 3}$ contains the
fractional coordinates, and $L\in\mathbb{R}^{d_L}$ denotes an
Euclidean lattice representation.
Given the chemical composition $A$ and an optional structural
condition $c$, the flow model (e.g., FlowMM~\cite{miller2024flowmm}) jointly generates the continuous
variables $(F,L)$ from sample distributions.

The lattice data $L_0$ are connected to Gaussian prior samples
$\epsilon_L\sim\mathcal{N}(0,I)$ through the linear path
$L_t=(1-t)L_0+t\epsilon_L$.
Starting from the prior at $t=1$, the corresponding reverse-time
generative ODE is discretized as:
\begin{equation*}
    L_{t-\Delta t}
    =
    L_t
    -
    \frac{
        \widehat{v}_{L,\theta}-L_t
    }{
        1-t
    }
    \Delta t .
    \label{eq:prelim_lattice_ode}
\end{equation*}

The fractional coordinates
$F\in\mathbb{T}^{N\times 3}$ lie on a product of flat tori.
Given a uniform prior sample
$F_1\sim\mathcal{U}([0,1)^{N\times3})$,
we compute the shortest periodic displacement as
$\Delta F=\operatorname{wrap}(F_1-F_0+0.5)-0.5$
and remove its mean translation to obtain
$v_F=\Delta F-\operatorname{Mean}(\Delta F)$.
Following FlowMM, inference uses the anti-annealed update controlled by $\gamma$ as:
\begin{equation*}
    F_{t-\Delta t}
    =
    \operatorname{wrap}
    \left[
        F_t+
        \bigl(1+\gamma(1-t)\bigr)
        \widehat{v}_{F,\theta}\Delta t
    \right].
    \label{eq:coordinate_generation}
\end{equation*}

\paragraph{RL on flow-matching models.}
The generation process can be formulated as a
finite-horizon Markov decision process $(\mathcal{S}, \mathcal{A}, \mathcal{P}, \mathcal{R})$ \cite{black2023training}.
At time $t$, the state is
$s_t=(c,t,x_t)$, where $c$ denotes the condition and
$x_t$ is the current latent structure.
The action $a_t=x_{t-\Delta t}$ specifies the next generated state,
and a terminal reward is assigned to the final structure $x_0$.
The policy objective is therefore:
\begin{equation*}
    J(\theta)
    =
    \mathbb{E}_{\tau\sim\pi_\theta}
    \left[
        R(x_0,c)
    \right],
    \qquad
    \tau=(x_T,\ldots,x_0),
    \label{eq:rl_objective}
\end{equation*}
which enables the direct optimization of physical and structural rewards without requiring differentiability.

\paragraph{ODE-to-SDE conversion.}
A pretrained flow model generates structures through the
deterministic probability-flow ODE:
$
    \mathrm{d}x_t
    =
    v_\theta(x_t,t,c)\,\mathrm{d}t,
    \label{eq:deterministic_flow}
$
whose transitions are Dirac distributions and therefore do not admit
tractable policy likelihoods.
Following Flow-GRPO~\cite{liu2025flowgrpo}, the ODE can be converted into an
associated stochastic process that preserves the pretrained
time-dependent marginals:
\begin{equation*}
    \begin{aligned}
x_{t+\Delta t} &= x_t + \left[ v_\theta (x_t, t) + \frac{\sigma_t^2}{2t} (x_t + (1 - t)v_\theta (x_t, t)) \right] \Delta t \\
&\quad + \sigma_t \sqrt{\Delta t} \boldsymbol{\epsilon}
\end{aligned}
\label{ode-sde}
\end{equation*}
where $\sigma_t = a\sqrt{\frac{t}{1-t}}$, $a$ is a scalar hyper-parameter to guide the noise level, and $\boldsymbol{\epsilon} \sim \mathcal{N}(0, \boldsymbol{I})$ injects stochasticity.
After discretization, the stochastic process yields Gaussian
transition probabilities, enabling tractable likelihood ratios and KL regularization for GRPO. Notably, we adopt this stochastic policy construction for GRPO and extend it to the coupled coordinate–lattice state of crystal flows in the next section.
\section{Method}

CrystalGRPO aligns reward-guided post-training with two requirements of budgeted CSP: identifying the reference polymorph and preserving its recoverability within a finite candidate set. To optimize these objectives with a pretrained flow backbone, we first convert the deterministic coordinate--lattice dynamics into a joint SDE policy over the complete continuous crystal state. On top of this joint SDE policy, CrystalGRPO combines dense energy-based feedback with a StructureMatcher-based recovery signal that identifies the reference structure during post-training.
To characterize and control the trade-off between reward-driven concentration and finite-budget target recovery, we instantiate two variants that share the joint policy and hybrid reward but differ in their degree of coverage preservation. \textbf{CrystalGRPO-Q} restricts reference-policy regularization to the early, high-noise portion of the trajectory, leaving later refinement steps more flexible and prioritizing single-draw recovery. \textbf{CrystalGRPO-C} extends reference regularization across the full trajectory and additionally applies a coverage-aware group advantage that protects valid but unmatched candidates from negative updates, thereby prioritizing Top-N target recovery.

\subsection{Joint Coordinate-Lattice SDE Policy}
\label{sec:sde}

Since the pretrained flow jointly evolves coordinates and lattices under a shared time parameterization, we directly stochasticize this joint continuous state, denoted as $z_t=(F_t,L_t)$. Let $\delta=1/T$, for $z\in\{F,L\}$, let $u^z_{\theta,t}$ denote the reverse-time velocity of field $z$ under the pretrained model. The transition mean can be formulated as:
\begin{equation*}
\begin{aligned}
    \mu^z_{\theta,t}
    =
    z_t+\delta
    \Bigg[
        u^z_{\theta,t}
        +
        \frac{\sigma_t^2}{2t}
        \Big(
            z_t+(1-t)u^z_{\theta,t}
        \Big)
    \Bigg].
\end{aligned}
\label{eq:joint_sde_mean}
\end{equation*}
The two fields share the same stochastic construction but retain
the field-specific drifts inherited from the pretrained flow. Their
geometric distinction enters the state update: the coordinate
proposal is wrapped onto the periodic torus, whereas the lattice
remains in its Euclidean representation. Specifically, the coordinate proposal is sampled in the Euclidean
covering space and subsequently wrapped into the periodic unit cell,
whereas the lattice update remains Euclidean: $\widetilde F_{t-\delta}
    =
    \mu^F_{\theta,t}
    +
    \sigma_t\sqrt{\delta}\,\epsilon^F_t$, $F_{t-\delta}
    =
    \operatorname{wrap}
    \bigl(\widetilde F_{t-\delta}\bigr)$, and $L_{t-\delta}
    =
    \mu^L_{\theta,t}
    +
    \sigma_t\sqrt{\delta}\,\epsilon^L_t$,
where the noises $\epsilon_t^F$ and $\epsilon_t^L$ are sampled independently
from standard Gaussian distributions. 


\paragraph{Size-normalized policy ratio.}
Following the Gaussian reparameterization used in
Flow-GRPO~\cite{liu2025flowgrpo}, the behavior and current transition
kernels share the same covariance $\sigma_t^2\delta I$. Their per-step
log-likelihood ratio therefore reduces to the difference between the
squared standardized residuals of the same realized proposal.

For $z\in\{F,L\}$, let $\epsilon_t^z$ denote the Gaussian noise sampled
during rollout under $\pi_{\mathrm{old}}$. The residual of the same
proposal under the current policy is obtained by reparameterization:
\begin{equation*}
    \epsilon_t^{z,\theta}
    =
    \epsilon_t^z
    +
    \frac{
        \mu_t^{z,\mathrm{old}}
        -
        \mu_t^{z,\theta}
    }{
        \sigma_t\sqrt{\delta}
    }.
\label{eq:current_policy_residual}
\end{equation*}
We regard the sampled Euclidean lift
$\widetilde F_{t-\delta}$ as the stochastic policy action, while
the wrapped coordinate $F_{t-\delta}$ is the next environment
state. The lifted proposal and its rollout noise are stored, so
the Gaussian policy score is evaluated on the same realized
action rather than on the many-to-one wrapped state alone. Besides, the joint Gaussian log-ratio scales with the number of generated
variables. We therefore use its Size-normalized form as:
\begin{equation*}
\begin{aligned}
    \widetilde{\ell}_t(\theta)
    &=
    \frac{1}{2N}
    \sum_{n=1}^{N}
    \left(
        \|\epsilon^F_{t,n}\|_2^2
        -
        \|\epsilon^{F,\theta}_{t,n}\|_2^2
    \right) 
    +
    \frac{1}{2d_L}
    \left(
        \|\epsilon^L_t\|_F^2
        -
        \|\epsilon^{L,\theta}_t\|_F^2
    \right), \\
    r_t(\theta)
    &=
    \exp\!\left[
        \operatorname{clip}
        \left(
            \widetilde{\ell}_t(\theta),
            -c_r,c_r
        \right)
    \right],
\end{aligned}
\label{eq:normalized_log_ratio}
\end{equation*}
where $c_r>0$ is a numerical clipping threshold applied before
exponentiation. The coordinate contribution is averaged over atoms, while the lattice contribution is averaged over its nine entries. This prevents the policy score from growing with crystal size or being dominated by the higher-dimensional coordinate field.

\paragraph{Size-normalized reference divergence.}
Let $\pi_{\mathrm{ref}}$ denote a frozen copy of the pretrained flow. At each integration step, the current and reference transition kernels share the same covariance $\sigma_t^2\delta I$ and differ only in their means. Their Gaussian KL divergence is therefore determined by the squared mean shift. As with the policy ratio, directly summing this shift over all
coordinates would make the divergence scale with crystal size and
would overweight the coordinate field relative to the lattice field.
We therefore define the field-wise normalized mean shifts as:
\begin{equation*}
\begin{aligned}
    d_t^F
    &=
    \frac{1}{N}
    \sum_{n=1}^{N}
    \left\|
        \mu_{t,n}^{F,\theta}
        -
        \mu_{t,n}^{F,\mathrm{ref}}
    \right\|_2^2,
    d_t^L
    =
    \frac{1}{d_L}
    \left\|
        \mu_t^{L,\theta}
        -
        \mu_t^{L,\mathrm{ref}}
    \right\|_F^2
\end{aligned}
\label{eq:normalized_mean_shift}
\end{equation*}
The resulting Size-normalized reference divergence is
$
    \widetilde D_{\mathrm{KL},t}
    =
    (d_t^F+d_t^L)/
         {2\sigma_t^2\delta}
$,
where the denominator follows from the KL divergence between Gaussian
transitions with identical covariance \cite{liu2025flowgrpo}. The normalization makes the
reference penalty comparable across crystal sizes and balances the
coordinate and lattice contributions.


\subsection{CSP-aligned Hybrid Reward}
\label{sec:reward}
CrystalGRPO combines a dense
energy-based term with a StructureMatcher-based recovery term, where the energy-based term is defined as the per-atom energy gap:
\begin{equation*}
\Delta E_g
= E(\widehat X_g)-E(X_{\mathrm{GT}}),
\label{eq:de}
\end{equation*}
where $E(\cdot)$ is the energy predicted by MACE \cite{batatia2025foundation} and we use a clip mechanism to make $\Delta E_g$ between $\in[-3,15]$. The upper clipping makes invalid or highly unphysical predictions receive an approximately constant worst-case reward, while the lower clipping prevents spuriously low energies caused by out-of-distribution extrapolation of the learned potential from dominating the policy update. We then define a target-recovery score using \texttt{StructureMatcher} \cite{ong2013python}:
\begin{equation*}
S_g=
\begin{cases}
\exp(-d_g/\tau), & \widehat X_g \text{ matches } X_{\mathrm{GT}},\\
0, & \text{otherwise},
\end{cases}
\label{eq:recovery_reward}
\end{equation*}
where $d_g$ is the RMS displacement after structural alignment and $\tau$
controls the sensitivity to geometric error. The final hybrid reward is:
\begin{equation*}
R_g=-\lambda_E\Delta E_g+\lambda_S S_g,
\label{eq:hybrid}
\end{equation*}
where $\lambda_E$ and $\lambda_S$ balance physical stability and target recovery, respectively. The energy term provides a dense ranking signal, while the structural term explicitly identifies the target polymorph and gives greater credit to more accurate matches.

\subsection{Coverage Preservation Across the GRPO Pipeline}
\label{sec:coverage}

The joint SDE policy and hybrid reward introduced above are shared by
both CrystalGRPO variants. \textbf{CrystalGRPO-Q} applies standard
group-relative optimization to these shared components. For a group of
$G$ candidates generated from the same condition \cite{shao2024deepseekmath}, its normalized
advantage is
$A_g=(R_g-\mu_{\mathcal G})/\sigma_{\mathcal G}$, where
$\mu_{\mathcal G}=\frac{1}{G}\sum_h R_h$.
By repeatedly reinforcing candidates with above-average rewards and
suppressing those below the group mean, CrystalGRPO-Q favors
single-draw recovery but may contract the multimodal distribution
needed for Top-N recovery.

To additionally preserve finite-budget target recoverability,
\textbf{CrystalGRPO-C} augments CrystalGRPO-Q with two complementary
mechanisms: a full trajectory KL constraint that limits cumulative
policy drift and a coverage-aware advantage that reduces destructive
within-group suppression.

\paragraph{Reference-policy KL constraint.}
Both CrystalGRPO variants regularize the learned policy toward a
frozen copy of the pretrained flow, but use different temporal
schedules. 
Using the normalized coordinate and lattice mean
shifts $d_t^F$ and $d_t^L$ defined above, we define:
\begin{equation*}
    \widetilde D_{\mathrm{KL},t}^{\,q}
    =
    w_t^{q}
    \frac{d_t^F+d_t^L}{2\sigma_t^2\delta},
    \qquad q\in\{\mathrm{Q},\mathrm{C}\}.
\label{eq:normalized_policy_kl}
\end{equation*}
Here, $w_t^{\mathrm{C}}=1$, whereas
$w_t^{\mathrm{Q}}=\mathbb{I}[t\geq\tau_{\mathrm{KL}}]$.
CrystalGRPO-C therefore applies reference regularization throughout the trajectory, while CrystalGRPO-Q applies it only during the early, high-noise steps that determine coarse structural basins. Removing the constraint at later low-noise steps allows greater flexibility for reward-guided structural refinement. 
\paragraph{Coverage-aware advantage.}
Besides the full trajectory KL, CrystalGRPO-C further modifies the group-relative advantage using the validity flag $v_g$ and the match flag $m_g=\mathbb{I}[S_g>0]$ induced by the recovery reward above. Specifically,
$\widetilde A_g=\max(A_g,0)$ when $v_g=1$ and $m_g=0$, and
$\widetilde A_g=A_g$ otherwise. For valid but unmatched candidates, this modification removes negative updates caused solely by falling below the group mean, without assigning any additional positive reward. Matched candidates retain their original advantages, so the continuous score $S_g=\exp(-d_g/\tau)$ continues to favor smaller RMS displacement, while invalid candidates remain suppressible. CrystalGRPO-Q and CrystalGRPO-C therefore share the same joint stochastic policy and target-aligned hybrid reward. CrystalGRPO-Q permits stronger reward-driven concentration, whereas CrystalGRPO-C combines full trajectory reference regularization with candidate-level protection to preserve target recoverability within a finite sampling budget.
\section{Discussion}
Recent analyses of RL post-training for language-model reasoning
have revealed a distinction between improving the probability of a
successful individual sample and expanding the set of solutions
recoverable under repeated sampling. \cite{yue2025does} observe that RL with verifiable rewards
substantially improves pass@$k$ at small $k$, especially pass@1,
while the corresponding base model can retain higher pass@$k$ at
large $k$. Their coverage analysis suggests that standard RL mainly
redistributes probability mass toward already recoverable rewarded
trajectories, improving sampling efficiency without necessarily
expanding the broader solution space. Relatedly, \cite{walder2025passk} argue that assigning rewards
independently to multiple rollouts primarily optimizes their
individual success probabilities, rather than the collective
best-of-$K$ utility of the candidate set.

A closely related objective mismatch arises in candidate-based CSP.
Standard group-relative optimization evaluates each generated
structure through its individual terminal reward. Candidates above
the group mean receive positive updates, whereas those below the mean
are suppressed. Repeated updates can therefore increase the
probability of already high-reward structural basins and improve
MR@1, while reducing the probability mass assigned to less frequent
basins that remain important for MR@20. This interpretation is
consistent with our composition-conditioned results: CrystalGRPO-Q
achieves stronger single-draw recovery, but its increased reward
concentration can reduce finite-budget target recovery.

CrystalGRPO-C addresses this mismatch through two complementary
forms of preservation. Full-trajectory reference regularization
limits cumulative redistribution away from the pretrained candidate
distribution, while the coverage-aware advantage prevents valid
unmatched candidates from receiving destructive negative updates
solely because their current rewards fall below the group mean.
Importantly, CrystalGRPO-C does not directly optimize MR@20 or
construct an unbiased estimator of best-of-$K$ success. Instead, it
provides a coverage-oriented surrogate that preserves target
recoverability under a fixed sampling budget. The connection to
pass@$K$ policy optimization is therefore conceptual: both recognize
that improving individual samples and preserving the collective
utility of a candidate set are distinct optimization objectives.
\section{Experiments}

\subsection{Experimental Setup}

We evaluate the two CrystalGRPO variants from four complementary
perspectives: comparison with existing CSP methods, consistency across pretrained backbones, component ablation, and PXRD-conditioned prediction. 


\paragraph{Datasets.}
Composition-conditioned CSP is evaluated on two standard benchmarks,
MP-20 and MPTS-52 \cite{xie2022crystal}. MP-20 contains
inorganic crystals with at most 20 atoms per unit cell, whereas MPTS-52
extends the evaluation to structures with up to 52 atoms per unit cell. PXRD-conditioned CSP is also evaluated on MP-20 by augmenting the composition with the target PXRD pattern \cite{li2025powder}, testing whether experimental conditioning can guide RL toward the target polymorph.

\paragraph{Baselines and backbones.}
The comparison includes representative CSP generators across major paradigms: CDVAE \cite{xie2022crystal}, DiffCSP \cite{jiao2023crystal}, CrystalFlow \cite{luo2025crystalflow}, PXRDGen \cite{li2025powder}. We chose PXRDGen and OMatG \cite{omatg} as our backbones. OMatG-IRL \cite{omatg-rl}, the most closely related RL-based CSP method, provides the RL baseline. 


\paragraph{Evaluation metrics.}
\label{eva_metric}
Complementary metrics capture both prediction quality and candidate-set
coverage. Top-1 match rate and RMSE measure single-sample quality, whereas
Top-20 counts a target as recovered if any of 20 candidates matches and reports
the lowest RMSE among the matched candidates. A prediction is counted as matched when Pymatgen's
\texttt{StructureMatcher} identifies it as equivalent to the ground-truth
structure under \texttt{stol}$=0.5$, \texttt{ltol}$=0.3$, and
\texttt{angle\_tol}$=10^\circ$~\cite{ong2013python}, and RMSE measures the atomic displacement between aligned predicted and ground-truth structures for matched cases. Throughout, match rates are computed over the full test set, with generation failures and structures failing the validity check counted as non-matches.

\begin{table*}[t]
\centering
\small
\setlength{\tabcolsep}{4pt}
\caption{Results for composition-conditioned CSP. PXRDGen and both CrystalGRPO variants use $N_t=200$ with velocity annealing on MP-20 and $N_t=100$ on MPTS-52. at inference, whereas CrystalGRPO incorporates a stochastic differential equation (SDE) during sampling. Values denoted with a dagger (\(\dag \)) are quoted directly from the original literature. \textbf{Bold} and \underline{underlined} values denote the \textbf{best} and \underline{second-best}.}
\label{tab:protocol-a}
\resizebox{0.9\linewidth}{!}{
\begin{tabular}{lcccccccc}
\toprule
\multirow{2}{*}{Method} & \multicolumn{4}{c}{MP-20} & \multicolumn{4}{c}{MPTS-52} \\
\cmidrule(lr){2-5}\cmidrule(lr){6-9}
 & MR@1 $\uparrow$ & RMSE@1 $\downarrow$ & MR@20 $\uparrow$ & RMSE@20 $\downarrow$
 & MR@1 $\uparrow$ & RMSE@1 $\downarrow$ & MR@20 $\uparrow$ & RMSE@20 $\downarrow$ \\
\midrule
CDVAE$^\dagger$        & 33.90 & 0.1045 & 66.95 & 0.1026 & 5.34  & 0.2106 & 20.79 & 0.2085\\
DiffCSP$^\dagger$      & 51.49 & \underline{0.0631} & 77.93 & 0.0492 & 12.19 & 0.1786 & 34.02 & 0.1749\\
CrystalFlow$^\dagger$   & 62.02  & 0.0710  & 78.34  & 0.0577  & 21.00    & 0.1613  & 37.81  & \underline{0.1584}\\

PXRDGen (backbone)  & 59.06 & 0.0683 & \underline{80.23} & 0.0540 & 18.67 & 0.1690 & \underline{38.74} & 0.1760\\
\midrule
CrystalGRPO-Q   & \textbf{64.63} & 0.0663 & 77.28 & \textbf{0.0484} & \textbf{23.04} & \underline{0.1535} & 37.51 & \textbf{0.1581}\\
CrystalGRPO-C             & \underline{62.20} & \textbf{0.0588} & \textbf{80.62} & \underline{0.0490} & \underline{22.49} & \textbf{0.1532} & \textbf{39.08} & 0.1618\\
\bottomrule
\end{tabular}
}
\end{table*}

\begin{table*}[t]
\centering
\small
\setlength{\tabcolsep}{3pt}
\caption{Result for cross-backbone RL Post-Training.
All settings use the backbone's default velocity annealing.
All RL methods use $N_t=50$ which is consistent with the setting in OMATG-IRL; Pretrained (native) uses each backbone's native inference step count
($N_t=200$ for PXRDGen on MP-20 and $N_t=210$ for OMatG on MP-20, $N_t=100$ for both backbones on MPTS-52).
Bold and underlined values denote the best and second-best results among
the $N_t=50$ settings.}
\label{tab:protocol-b}
\resizebox{0.8\linewidth}{!}{
\begin{tabular}{llcccccccc}
\toprule
\multirow{2}{*}{Method}
& \multirow{2}{*}{Sampling}
& \multicolumn{4}{c}{MP-20}
& \multicolumn{4}{c}{MPTS-52} \\
\cmidrule(lr){3-6}\cmidrule(lr){7-10}
&
& MR@1 
& RMSE@1 
& MR@20 
& RMSE@20
& MR@1
& RMSE@1
& MR@20 
& RMSE@20 \\
\midrule
\multicolumn{10}{l}{\textit{backbone}} \\
PXRDGen
& ODE / native
& 59.06 & 0.0683 & 80.23 & 0.0540 & 18.67 & 0.1690 & 38.74 & 0.1760 \\

PXRDGen
& ODE / 50

& 59.43 & 0.1026 & \textbf{80.52} & 0.0839 
& 19.01 & 0.2090 & \underline{38.96} & 0.2045\\
PXRDGen
& SDE / 50
& 59.29 & 0.1837 & 80.29 & 0.1476
& 18.38 & 0.2342 & 37.80 & 0.2294 \\
\cmidrule(lr){1-10}
OMatG-IRL
& SDE / 50
& \underline{60.82} & 0.1096 & 78.52 & 0.0922
& 18.38 & 0.1884 & 38.60 & 0.1951 \\
CrystalGRPO-Q
& SDE / 50

& \textbf{63.70}
& \underline{0.0789}
& 78.08
& \textbf{0.0497}
& \textbf{22.31}
& \textbf{0.1687}
& 38.18
& \textbf{0.1722}\\
CrystalGRPO-C
& SDE / 50
& 60.61
& \textbf{0.0702}
& \underline{80.41}
& \underline{0.0520}
& \underline{21.46} & \underline{0.1777} & \textbf{39.00} & \underline{0.1751}\\
\midrule
\midrule
\multicolumn{10}{l}{\textit{backbone}} \\
OMatG
& ODE / native

& 63.55 & 0.0697 & 74.62 & 0.0617
& 25.25 & 0.1965 & 36.28 & 0.1788
 \\
OMatG
& ODE / 50

& 63.59 & 0.1249 & 74.76 & 0.1036 
& 22.70 & 0.2953 & 34.17 & 0.2726\\
OMatG
& SDE / 50
& 63.43 & 0.2060 & 74.22 & 0.1771
& 22.36 & 0.2975 & 33.87 & 0.2712 \\
\cmidrule(lr){1-10}
OMatG-IRL
& SDE / 50

& \underline{64.28} & 0.0903 & 74.33 & 0.0780
& \underline{25.30} & 0.1721 & 36.15 & 0.1630\\
CrystalGRPO-Q
& SDE / 50
& \textbf{64.32}
& \textbf{0.0722}
& \underline{76.46}
& \underline{0.0487} 
& \textbf{26.57}
& \underline{0.1603}
& \underline{36.85}
& \textbf{0.1497}\\
CrystalGRPO-C
& SDE / 50
& 63.81
& \underline{0.0776}
& \textbf{77.72}
& \textbf{0.0484} 
& 25.26 & 0.1656 & \textbf{37.36} & \underline{0.1501}\\
\bottomrule
\end{tabular}
}
\end{table*}

\subsection{Comparison with Existing CSP Methods}
\label{sec:main_comparison}

We first compare CrystalGRPO with existing methods under the standard composition-conditioned setting. Both variants are initialized from PXRDGen and use composition alone as input. As shown in Table~\ref{tab:protocol-a}, \textbf{CrystalGRPO-Q} achieves the highest MR@1 on both datasets. On MP-20, it improves MR@1 from 59.06\% to 64.63\%, and on MPTS-52 from 18.67\% to 23.04\%. The corresponding MR@20 values, however, decrease from 80.23\% to 77.28\% and from 38.74\% to 37.51\%, respectively. This consistent pattern exposes the quality--coverage trade-off produced by reward-driven optimization: stronger single-sample recovery can be accompanied by contraction of the candidate distribution. \textbf{CrystalGRPO-C} shifts the operating point toward finite-budget recovery. It achieves MR@1/MR@20 values of 62.20\%/80.62\% on MP-20 and 22.49\%/39.08\% on MPTS-52, giving the highest MR@20 on both datasets while retaining clear MR@1 improvements over the pretrained backbone. Both variants also improve geometric accuracy. CrystalGRPO-Q reduces RMSE@20 from 0.0540 to 0.0484 on MP-20 and from 0.1760 to 0.1581 on MPTS-52, while CrystalGRPO-C reduces RMSE@1 to 0.0588 and 0.1532, respectively. These results support the finite-budget formulation introduced earlier: the shared joint policy and hybrid reward enable effective target-oriented post-training, while full-trajectory reference regularization and the coverage-aware advantage allow CrystalGRPO-C to preserve candidate-set recovery without discarding the single-sample gains obtained through RL.

\subsection{Cross-Backbone RL Post-Training}
\label{sec:backbone_generalization}

Table~\ref{tab:protocol-b} evaluates CrystalGRPO on two pretrained
backbones with different flow constructions. The ODE/50 and SDE/50 controls for base model first isolate the effect of stochasticization before any policy update. Across the four backbone--dataset settings, replacing deterministic ODE sampling with the SDE process changes MR@1 by at most 0.63 percentage points and MR@20 by at most 1.16 points. The corresponding RMSE values increase in every setting, indicating that the SDE process largely preserves which structural basins remain recoverable from the pretrained backbone, although the injected noise reduces within-basin geometric precision. 
The two variants exhibit consistent operating behavior across both
backbones. Relative to the matched ODE/50 controls,
CrystalGRPO-Q raises MR@1 from 59.43\% to 63.70\% on MP-20 and from 19.01\% to 22.31\% on MPTS-52 with PXRDGen. With OMatG, it raises MR@1 from 63.59\% to 64.32\% and from 22.70\% to 26.57\%,
respectively. CrystalGRPO-C gives the strongest MR@20 among the RL
methods in all four settings, reaching 80.41\% and 39.00\% with
PXRDGen, and 77.72\% and 37.36\% with OMatG. On MP-20 with PXRDGen, its MR@20 remains essentially matched to the ODE/50 control; in the other three settings, it exceeds the matched deterministic baseline. This repeated pattern supports CrystalGRPO-Q as the single-draw operating point and CrystalGRPO-C as the coverage-preserving operating point for finite-budget recovery. 

Table~2 provides an end-to-end comparison with OMatG-IRL.
In the evaluated OMatG-IRL configuration, policy optimization is
applied only to the fractional coordinates, while the lattice
continues to follow the frozen pretrained dynamics. CrystalGRPO
instead optimizes both coordinate and lattice transitions.
Across all four backbone--dataset settings, both CrystalGRPO
variants achieve lower RMSE@1 and RMSE@20 than OMatG-IRL.
CrystalGRPO-Q additionally achieves higher MR@1 in all four
settings, while CrystalGRPO-C achieves higher MR@20 in all four.
The improvements also transfer across two backbones with
substantially different lattice generation schemes. OMatG
initializes the lattice from a dataset-informed prior, whereas
PXRDGen starts from an unstructured Gaussian distribution over
raw $3\times3$ lattice matrices. CrystalGRPO improves recovery
under both settings, suggesting that the complete framework is
not tied to a particular lattice initialization scheme.

\begin{table}
\centering
\caption{Ablation on MP-20. Each step inherits the preceding setting and additionally removes the named component. All variants use $N_t{=}50$, matching the sampling setup in Table~\ref{tab:protocol-b}. Bold and underlined values indicate the best and second-best available results.}
\label{tab:coverage_ablation}
\resultstableformat
\begin{tabular}{@{}lcccc@{}}
\toprule
Variant & \mronehead & \rmseonehead & \mrtwentyhead & \rmsetwentyhead \\
\midrule
CrystalGRPO-C & 60.61 & \textbf{0.0702} & \textbf{80.41} & 0.0520\\
Full $\rightarrow$ early-stage KL & 63.02 & \underline{0.0783} & 79.31 & 0.0516 \\
w/o Coverage & \textbf{63.70} & 0.0789 & 78.08 & 0.0497 \\
w/o Hybrid Reward & 62.89 & 0.0854 & 76.10 & \textbf{0.0444} \\
w/o Joint SDE & \underline{63.32} & 0.1037 & 74.70 & \underline{0.0447} \\
\midrule
PXRDGen (SDE/50) & 59.29 & 0.1837 & \underline{80.29} & 0.1476  \\
\bottomrule
\end{tabular}
\end{table}

\begin{table}
\centering
\caption{Best-of-20 RMSE on shared matches and targets additionally
recovered by CrystalGRPO-C.}
\label{tab:coverage_recovery}
\small
\setlength{\tabcolsep}{4pt}
\resizebox{0.42\textwidth}{!}{
\begin{tabular}{@{}lccc@{}}
\toprule
Target subset
& Count
& \shortstack{CrystalGRPO-Q\\RMSE}
& \shortstack{CrystalGRPO-C\\RMSE} \\
\midrule
Shared matches & 6,940 & 0.0452 & 0.0446 \\
CrystalGRPO-C only & 334 & -- & 0.2192 \\
\bottomrule
\end{tabular}
}
\end{table}

\subsection{Ablation Study}
\label{sec:ablation}

Table~\ref{tab:coverage_ablation} traces the transition from the
coverage-preserving CrystalGRPO-C to the quality-oriented
CrystalGRPO-Q, and then cumulatively removes the shared reward and
policy components. Replacing the full-trajectory reference constraint of CrystalGRPO-C with early-stage regularization increases MR@1 from 60.61\% to 63.02\%, but reduces MR@20 from 80.41\% to 79.31\%. Removing the coverage-aware advantage then yields CrystalGRPO-Q, further increasing MR@1 to 63.70\% while decreasing MR@20 to 78.08\%. These results show how the two coverage mechanisms shift the operating point: relaxing reference anchoring and within-group protection favors single-draw recovery, whereas their combination preserves the candidate distribution needed for finite-budget recovery. 
Although CrystalGRPO-Q obtains a lower RMSE@20 of 0.0497, this does not indicate better overall recovery. RMSE is averaged only over successfully matched structures, and the lower match rate leaves a smaller, easier subset of targets in the evaluation. As shown in Table ~\ref{tab:coverage_recovery}, the modestly higher RMSE under coverage preservation is driven mainly by newly recovered, lower-fidelity matches rather than by degraded refinement of targets that were already recoverable. Coverage preservation therefore extends recovery to more difficult, low-frequency targets, at the cost of a small increase in RMSE averaged over successful matches.
Furthermore, replacing the hybrid reward with energy alone reduces MR@1 from 63.70\% to 62.89\% and MR@20 from 78.08\% to 76.10\%. The larger drop in candidate-set recovery is consistent with our motivation: energy provides useful physical ranking, but cannot reliably identify the reference polymorph among competing low-energy structures. 
The final cumulative configuration additionally removes the joint coordinate--lattice SDE and reinforces only the coordinate field while leaving the lattice under the frozen pretrained process. MR@1 remains similar at 63.32\%, but RMSE@1 worsens from 0.0854 to 0.1037 and MR@20 falls further to 74.70\%.

\begin{table}
\centering
\caption{PXRD-guided CSP on MP-20. All methods are conditioned on composition and the target PXRD pattern. Both variants use energy-only reward during RL. All settings use the backbone's default velocity annealing with $N_t=200$.} 
\label{tab:pxrd_guided} 
\resultstableformat 
\begin{tabular}{@{}lcccc@{}} 
\toprule Method & \mronehead & \rmseonehead & \mrtwentyhead & \rmsetwentyhead \\ 
\midrule PXRDGen (base) & 68.41 & 0.0705 & \underline{84.73} & 0.0484 \\ 
CrystalGRPO-Q & \textbf{72.71} & \textbf{0.0634} & 84.13 & \textbf{0.0439} \\ 
CrystalGRPO-C & \underline{69.50} & \underline{0.0679} & \textbf{85.39} & 0.0500 \\ 
\bottomrule 
\end{tabular} 
\end{table}


\begin{table}
  \centering
  \footnotesize
  \setlength{\tabcolsep}{4pt}
  \caption{ Low-energy GT-mismatched candidates: valid, GT-mismatched
    structures within $\Delta E\!\le\!0.1$\,eV/atom of GT. All counts exact,
    $N{=}9046$. Both Models use energy-only reward during RL training with $N_t=200$ during inference.}
  \label{tab:poly}
  \begin{tabular}{@{}l cc c cc@{}}
    \toprule
    & \multicolumn{2}{c}{Match\%\,$\uparrow$} && \multicolumn{2}{c}{Low-E mismatched candidates\,$\downarrow$} \\
    \cmidrule(lr){2-3}\cmidrule(lr){5-6}
    Model & base & +RL && base & +RL \\
    \midrule
    w/o XRD & 59.06 & 63.60 && 592 & 1562~{\scriptsize($+164\%$)} \\
    w/\,XRD & 68.41 & \textbf{72.71} && 812 & \textbf{810}~{\scriptsize($-0\%$)} \\
    \bottomrule
  \end{tabular}
\end{table}



\subsection{PXRD-Guided Crystal Structure Prediction}
\label{sec:pxrd_guided}

We next examine how target-specific conditioning changes the
interaction between energy guidance and candidate coverage. In this setting the reward is energy-only, both CrystalGRPO variants are conditioned on the target PXRD pattern in addition to composition; PXRD is used only as an input condition \cite{li2025powder} and the reward is purely based on energy instead of hybrid reward. As shown in
Table~\ref{tab:pxrd_guided}, \textbf{CrystalGRPO-Q} achieves the
highest MR@1, improving the PXRDGen backbone from 68.41\% to 72.71\%. Its MR@20 decreases only slightly, from 84.73\% to 84.13\%, compared with the substantially larger coverage loss observed under composition-only conditioning. \textbf{CrystalGRPO-C} instead reaches the highest MR@20 of 85.39\%, while retaining an MR@1 improvement to 69.50\%. Target-specific conditioning therefore weakens, but does not entirely remove, the quality--coverage trade-off between the two operating modes. Table~\ref{tab:poly} isolates the role of target information using
controlled energy-only CrystalGRPO-Q runs. Without PXRD, RL raises the match rate from 59.06\% to 63.60\%, but the number of valid, reference-mismatched candidates within 0.1\,eV/atom of the reference increases from 592 to 1562, a 164\% increase. With PXRD conditioning, the match rate improves from 68.41\% to 72.71\%, while the number of low-energy mismatched candidates remains essentially unchanged, from 812 to 810. These results support the distinction developed in the Introduction: predicted energy supplies useful physical ranking but does not by itself identify the reference structure. PXRD narrows the generation toward a target-specific structural basin, making energy a more reliable refinement signal and reducing the distributional ambiguity
that coverage preservation must control.

\section{Conclusion}
We presented CrystalGRPO, a CSP-aligned post-training framework for
flow-based crystal generators. It extends existing ODE-to-SDE policy
constructions to the joint coordinate--lattice state, combines
MACE-predicted energy with a recovery score that provides the target
identity absent from energy alone, and exposes the quality--coverage
trade-off through two operating modes. Across MP-20 and MPTS-52 with
PXRDGen and OMatG backbones, both variants reduce RMSE@1 and RMSE@20
relative to coordinate-only reinforcement, while CrystalGRPO-Q
consistently improves MR@1 and CrystalGRPO-C achieves higher MR@20.
These results demonstrate a flexible post-training framework for
adapting pretrained crystal generators to different finite-budget
recovery objectives. Future work will
explore larger and more complex systems, particularly organic
molecular crystals with richer conformational variation.

\section*{Acknowledgments}

This research was sponsored by the U.S. Department of Energy, Office of Science, Office of Basic Energy Sciences, and the Established Program to Stimulate Competitive Research (EPSCoR) under the DOE Early Career Award No. DE-SC0024866. The computing resources are provided by ACCESS (TG-MAT230046).

{
    \small
    \bibliographystyle{ieeenat_fullname}
    \bibliography{main}
}

\clearpage
\setcounter{page}{1}
\maketitlesupplementary

\section{Detailed Preliminaries and Derivations}

\subsection{Crystal Structure Prediction} Crystal structure prediction aims to recover the periodic atomic
arrangement and lattice consistent with a given chemical composition
and, when available, experimental observations. A crystal is
represented as $M=(A,F,L)$, where $A$ denotes the atomic species,
$F$ the fractional coordinates, and $L$ the lattice matrix.


\subsection{Flow-Based crystal strucature prediction}

In this work, we apply flow matching models as our backbone to jointly generate the lattice matrices ($L$) and fractional coordinates ($F$) of crystal structures conditioned on chemical composition ($A$) or PXRD pattern. Here we will provide a detailed introduction about the formulation and network structure of PXRDGen \cite{li2025powder}.
\subsubsection{Flow on Lattice}
As the lattices are defined in the Euclidean space $\mathbb{R}^{3 \times 3}$, the flow path from data $\mathbf{L}_0$ to prior noise $\mathbf{L}_1 \sim \mathcal{N}(\mathbf{0}, \mathbf{I})$ can be simply constructed as
\begin{equation*}
    q(\mathbf{L}_t | \mathbf{L}_0) = \mathcal{N}(\mathbf{L}_t | (1 - t)\mathbf{L}_0, t\mathbf{I}).
\end{equation*}

The training objective is defined to predict the terminal noise state $\mathbf{L}_1$ (denoted as $\boldsymbol{\epsilon}_{\mathbf{L}}$ for clarity) to ensure numerical stability during lattice deformation:
\begin{equation*}
    \mathcal{L}_{\mathbf{L}, \text{flow}} = \mathbb{E}_{\boldsymbol{\epsilon}_{\mathbf{L}} , t } \left[ \left\| \boldsymbol{\epsilon}_{\mathbf{L}} - \hat{\boldsymbol{\epsilon}}_{\mathbf{L}}(\mathcal{M}_t, t, c) \right\|_2^2 \right].
\end{equation*}
where $\epsilon_{\mathbf{L}} \sim \mathcal{N}(\mathbf{0}, \mathbf{I}), t \sim \mathcal{U}(0, 1)$. The generative ODE process follows:
\begin{equation*}
    \mathbf{L}_{t-\Delta t} = \mathbf{L}_t - \frac{\hat{\boldsymbol{\epsilon}}_{\mathbf{L}} - \mathbf{L}_t}{1 - t} \Delta t.
\end{equation*}

\subsubsection{Flow on Fractional Coordinate}
The fractional coordinates $\mathbf{F}$ of an $N$-atom point cloud inherently live on an $N \times 3$-dimensional product of flat tori $\mathbb{T}^{N \times 3}$, constrained by periodic boundary conditions. To construct a valid flow matching path on this periodic manifold, we utilize geodesic paths that seamlessly cross the periodic boundaries. We first formulate the exponential and logarithmic maps on the torus for a given atom index $i$:
\begin{equation*}
    \begin{aligned}
        \exp_{\mathbf{F}^i}(\mathbf{V}^i) := \mathbf{F}^i + \mathbf{V}^i - \lfloor \mathbf{X}^i + \mathbf{V}^i \rfloor, \\
        \quad \log_{\mathbf{F}_0^i}(\mathbf{F}_1^i) := \frac{1}{2\pi} \text{atan2} \left[ \sin(\boldsymbol{\omega}^i), \cos(\boldsymbol{\omega}^i) \right],
    \end{aligned}
\end{equation*}
where $\boldsymbol{\omega}^i := 2\pi(\mathbf{F}_1^i - \mathbf{F}_0^i)$ and $\mathbf{V}$ represents a tangent velocity vector. 

To rigorously address the global translation-invariance, we acquire the final target velocity field by removing the mean torus translation from the base logarithmic map:
\begin{equation*}
    \boldsymbol{v}_{\mathbf{X}}(\mathbf{F}_0, \mathbf{F}_1) := \log_{\mathbf{F}_0}(\mathbf{F}_1) - \frac{1}{N} \sum_{i=1}^N \log_{\mathbf{F}_0^i}(\mathbf{F}_1^i)
\end{equation*}

The time-dependent state along the geodesic path is elegantly constructed as $\mathbf{F}_t = \exp_{\mathbf{F}_0}(t \boldsymbol{v}_{\mathbf{F}})$. The flow matching training objective on coordinates is optimized to regress this invariant vector field:
\begin{equation*}
    \mathcal{L}_{\mathbf{F}, \text{flow}} = \mathbb{E}_{t , \mathbf{F}_1 } \left[ \left\| \hat{\boldsymbol{v}}_{\mathbf{F}}(\mathcal{M}_t, t, c) - \boldsymbol{v}_{\mathbf{F}}(\mathbf{F}_0, \mathbf{F}_1) \right\|_2^2 \right]
\end{equation*}
where $t \sim \mathcal{U}(0,1), \mathbf{F}_1 \sim \mathcal{U}(0,1)$. To accelerate the structural relaxation into deep energy basins during inference, we apply an anti-annealing strategy. Utilizing the defined exponential map to respect the boundary constraints, the generative ODE is discretized as:
\begin{equation*}
    \mathbf{F}_{t-\Delta t} = \exp_{\mathbf{F}_t} \left( \big(1 + \gamma(1 - t)\big) \hat{\boldsymbol{v}}_{\mathbf{F}} \Delta t \right)
\end{equation*}

\subsubsection{Decoder Network}
 A graph neural network serves as the backbone decoder to jointly model $\mathbf{L}$ and $\mathbf{F}$ \cite{jiao2023crystal}. It takes atom types $f_{\text{atom}}(\mathbf{A}_i)$, time embeddings $f_{\text{pos}}(t)$, and condition features $c$ (e.g., $f_{\text{XRD}}$ encoded from the PXRD pattern) as inputs, where the PXRD encoder is a MAR model. The initial atom representation is acquired via a multi-layer perceptron (MLP) as $h_i^{(0)} = \rho \left( f_{\text{atom}}(\mathbf{A}_i), f_{\text{pos}}(t), c \right)$. 

The message passing mechanism is modeled as:
\begin{equation*}
    m_{ij}^{(s)} = \phi_m \left( h_i^{(s-1)}, h_j^{(s-1)}, \mathbf{L}^\top \mathbf{L}, \psi(\mathbf{F}_j - \mathbf{F}_i) \right),
\end{equation*}
\begin{equation*}
    m_i^{(s)} = \sum_{j=1}^N m_{ij}^{(s)},
\end{equation*}
\begin{equation*}
    h_i^{(s)} = h_i^{(s-1)} + \phi_h \left( h_i^{(s-1)} + m_i^{(s)} \right),
\end{equation*}
where $\phi_m, \phi_h$ are MLPs, and $\psi$ represents periodic radial basis embeddings. For the final output layer, the network predicts the lattice noise and the coordinate velocity:
\begin{equation*}
    \hat{\boldsymbol{\epsilon}}_{\mathbf{L}} = \mathbf{L} \phi_{\mathbf{L}} \left( \frac{1}{N} \sum_{i=1}^N h_i^{(S)} \right),
\end{equation*}
\begin{equation*}
    \hat{\boldsymbol{v}}_{\mathbf{F}, i} = \phi_{\mathbf{F}} \left( h_i^{(S)} \right).
\end{equation*}
Crucially, the predicted velocity $\hat{\boldsymbol{v}}_{\mathbf{F}}$ is further subjected to a de-translation operation $\hat{\boldsymbol{v}}_{\mathbf{F}} \leftarrow \hat{\boldsymbol{v}}_{\mathbf{F}} - \text{Mean}(\hat{\boldsymbol{v}}_{\mathbf{F}})$ to ensure that the vector field perfectly adheres to the translation-invariant manifold.

\subsection{PXRD-Conditioned Base Model and MAE Encoder}
\label{sec:pxrd_conditioned_base}

\paragraph{PXRD representation.}
For the PXRD-conditioned backbone, each crystal is associated with a
simulated powder diffraction pattern covering the $2\theta$ range from
$5^\circ$ to $80^\circ$ with an interval of $0.01^\circ$. The resulting
intensity profile contains 7,500 values and is normalized by its maximum
intensity. PXRD patterns are simulated using GSAS-II with a Cu X-ray
instrument profile. The chemical composition, atom types, and number of
atoms are provided as part of the CSP input; the model predicts the
fractional coordinates and lattice matrix conditioned on the PXRD pattern.

\paragraph{MAE-based PXRD encoder.}
We encode the one-dimensional PXRD intensity profile using a masked
autoencoder. Given a PXRD pattern
$\mathbf{p}\in\mathbb{R}^{7500}$, we divide it into 750 non-overlapping
patches of length 10:
\begin{equation}
    \mathbf{p}
    =
    \left[
    \mathbf{p}_1,\ldots,\mathbf{p}_{750}
    \right],
    \qquad
    \mathbf{p}_j\in\mathbb{R}^{10}.
\end{equation}
Each patch is projected to a 128-dimensional token using a linear layer.
Fixed one-dimensional sine--cosine positional embeddings are added to the
patch tokens, together with a learnable \texttt{[CLS]} token.

During encoder pretraining, 40\% of the patch tokens are randomly masked.
The visible tokens are processed by four Transformer blocks, each containing
four attention heads and an MLP with an expansion ratio of four. The
Transformer output corresponding to the \texttt{[CLS]} token is projected
from 128 to 256 dimensions to produce the global PXRD representation
$\mathbf{z}^{\mathrm{xrd}}\in\mathbb{R}^{256}$.

A two-layer Transformer decoder reconstructs the masked patches. Let
$\mathcal{M}$ denote the set of masked patches. The masked reconstruction
loss is
\begin{equation}
    \mathcal{L}_{\mathrm{MAE}}
    =
    \frac{1}{|\mathcal{M}|}
    \sum_{j\in\mathcal{M}}
    \frac{1}{10}
    \left\|
    \widehat{\mathbf{p}}_j-\mathbf{p}_j
    \right\|_2^2.
\end{equation}
Only masked patches contribute to this reconstruction objective.

In addition to masked reconstruction, the PXRD encoder is aligned with a
structure encoder using symmetric contrastive learning. The structure
encoder maps the corresponding ground-truth crystal to a
256-dimensional representation $\mathbf{z}^{\mathrm{str}}$. After
$\ell_2$ normalization, the cross-modal similarity between the $i$-th PXRD
pattern and the $j$-th structure is
\begin{equation}
    s_{ij}
    =
    \frac{
    \left\langle
    \overline{\mathbf{z}}^{\mathrm{xrd}}_i,
    \overline{\mathbf{z}}^{\mathrm{str}}_j
    \right\rangle
    }{T_c},
\end{equation}
where $T_c=0.05$ is the contrastive temperature. The bidirectional
contrastive loss is
\begin{equation}
    \mathcal{L}_{\mathrm{CL}}
    =
    \frac{1}{2}
    \left[
    \operatorname{CE}(\mathbf{S},\mathbf{I})
    +
    \operatorname{CE}(\mathbf{S}^{\mathsf T},\mathbf{I})
    \right],
\end{equation}
where the diagonal entries represent the matched PXRD--structure pairs in
the same mini-batch. The complete encoder-pretraining objective is
\begin{equation}
    \mathcal{L}_{\mathrm{enc}}
    =
    \mathcal{L}_{\mathrm{CL}}
    +
    0.5\,\mathcal{L}_{\mathrm{MAE}}.
\end{equation}
The structure encoder is used only during representation pretraining and is
not required by the subsequent generative model.

After pretraining, we set the masking ratio to zero and use the complete
PXRD profile to compute the condition embedding. The MAE decoder is bypassed,
and the PXRD encoder is frozen during both flow-matching pretraining and
CrystalGRPO optimization. The resulting condition is normalized as
\begin{equation}
    \mathbf{c}_{\mathrm{xrd}}
    =
    \frac{\mathbf{z}^{\mathrm{xrd}}}
    {\|\mathbf{z}^{\mathrm{xrd}}\|_2}.
\end{equation}

\paragraph{PXRD-conditioned flow model.}
The generative backbone is a conditional crystal flow model operating
jointly on fractional coordinates and lattice matrices. The conditional CSP decoder receives the diffusion-time embedding, atom-type
embedding, current fractional coordinates, current lattice, and the frozen PXRD condition $\mathbf{c}_{\mathrm{xrd}}$. The PXRD and time embeddings are
broadcast to all atoms of the corresponding crystal and concatenated with
the atom embeddings:
\begin{equation}
    \mathbf{h}^{(0)}_i
    =
    \operatorname{MLP}
    \left(
    \left[
    \operatorname{Emb}(a_i),
    \phi(t),
    \mathbf{c}_{\mathrm{xrd}}
    \right]
    \right).
\end{equation}
The decoder contains six fully connected crystal-graph message-passing
layers with a hidden dimension of 512. Edge features include the two endpoint
node representations, a sinusoidal encoding of the periodic fractional
displacement, and the lattice metric
$\mathbf{L}_t\mathbf{L}_t^{\mathsf T}$. The network produces a per-atom
coordinate field and a graph-level lattice prediction obtained by mean
pooling the node features, while all other decoder settings remain unchanged.

\subsection{GRPO Framework}
The goal of Reinforcement Learning is to learn a policy that maximizes expected cumulative reward. GRPO achieves this by optimizing its policy model to maximize the following objective:
\begin{equation*}
    \mathcal{J}_{\text{Flow-GRPO}}(\theta) = \mathbb{E}_{\boldsymbol{c} \sim \mathcal{C}, \{\boldsymbol{x}^i\}_{i=1}^G \sim \pi_{\theta_{\text{old}}}(\cdot \mid \boldsymbol{c})} f(r, \hat{A}, \theta, \epsilon, \beta)
\end{equation*}
where the $f(r, \hat{A}, \theta, \epsilon, \beta)$ is defined as:
\begin{equation*}
    \begin{aligned} f(r, \hat{A}, \theta, \epsilon, \beta) &= \frac{1}{G} \sum_{i=1}^G \frac{1}{T} \sum_{t=0}^{T-1} \Big( \min \Big( r_t^i(\theta) \hat{A}_t^i, \\ &\quad \text{clip}\left( r_t^i(\theta), 1 - \epsilon, 1 + \epsilon \right) \hat{A}_t^i \Big)  \\ &\quad - \beta D_{\text{KL}} \left( \pi_\theta \middle\| \pi_{\text{ref}} \right)\Big)
    \end{aligned}
\end{equation*}
\begin{equation*}
    r_t^i(\theta) = \frac{p_\theta \left( \boldsymbol{x}_{t-1}^i \mid \boldsymbol{x}_t^i, \boldsymbol{c} \right)}{p_{\theta_{\text{old}}} \left( \boldsymbol{x}_{t-1}^i \mid \boldsymbol{x}_t^i, \boldsymbol{c} \right)}
\end{equation*}
Given a condition $\boldsymbol{c}$, the flow model $p_\theta$ samples a group of $G$ individual crystal structures $\left\{ \boldsymbol{x}_0^i \right\}_{i=1}^G$ and the corresponding reverse-time trajectories $\left\{ \left( \boldsymbol{x}_T^i, \boldsymbol{x}_{T-1}^i, \dots, \boldsymbol{x}_0^i \right) \right\}_{i=1}^G$, where $T$ is the number of integration steps. Then, the advantage of the $i$-th crystal is calculated by normalizing the group-level rewards as follows:
\begin{equation*}
    \hat{A}_t^i = \frac{R\left( \boldsymbol{x}_0^i, \boldsymbol{c} \right) - \text{mean} \left( \left\{ R\left( \boldsymbol{x}_0^i, \boldsymbol{c} \right) \right\}_{i=1}^G \right)}{\text{std} \left( \left\{ R\left( \boldsymbol{x}_0^i, \boldsymbol{c} \right) \right\}_{i=1}^G \right)}
\end{equation*}

\subsection{RL on Flow Matching Model} Within reinforcement learning, a sequential decision-making problem can be formulated as a Markov decision process (MDP) $(\mathcal{S}, \mathcal{A}, \rho_{0}, \mathcal{P}, \mathcal{R})$, where $\mathcal{S}$ denotes the state space, $\mathcal{A}$ presents the action space, $\rho_{0}$ denotes the distribution of initial states, $\mathcal{P}$ is the transition kernel, and $\mathcal{R}$ is the reward function. At each timestep t with a state $\mathbf{s}_t \in \mathcal{S}$ the agent takes an action $a_t \in \mathcal{A}$ according to a policy $\pi(a|s)$ and thereby receives a reward $\mathcal{R}(\mathbf{s}_t, \mathbf{a}_t)$, moving to a new state based on the transition function  $ \mathcal{P}(\mathbf{s}_{t+1} |\mathbf{s}_t, \mathbf{a}_t)$.
The iterative denoising process in flow matching models can be formulated as a Markov game, where the state at step $t$ is $s_t \triangleq (c, t, \boldsymbol{x}_t)$, the action is the denoised sample $a_t \triangleq \boldsymbol{x}_{t-1}$ predicted by the model, and the policy is $\pi(a_t \mid s_t) \triangleq p_\theta(\boldsymbol{x}_{t-1} \mid \boldsymbol{x}_t, c)$. The transition is deterministic: $P(s_{t+1} \mid s_t, a_t) \triangleq (\delta_c, \delta_{t-1}, \delta_{\boldsymbol{x}_{t-1}})$, and the initial state distribution is $\rho_0(s_0) \triangleq (p(c), \delta_T, \mathcal{N}(\mathbf{0}, \mathbf{I}))$, where $\delta_y$ is the Dirac delta distribution centered at $y$. The reward is only given at the final step: $R(s_t, a_t) \triangleq r(\boldsymbol{x}_0, c)$ if $t = 0$, and $0$ otherwise.

\subsection{From ODE to SDE Sampler}
Typically, flow matching models predict the velocity $v_t$ and use a deterministic ODE for the denoising process:
\begin{equation*}
    dx_t = v_tdt
\end{equation*}
However, the deterministic ordinary Differential equation (ODE) sampling inherent to flow models fundamentally fails to satisfy two critical requirements of GRPO. First, GRPO requires stochastic rollouts to generate distinct trajectories for exploration and group-relative advantage estimation. Second, policy optimization requires tractable step-wise transition probabilities to compute importance ratios, which are unavailable for deterministic Dirac transitions.
To this end, Flow-GRPO \cite{liu2025flowgrpo} injects additional noise to sampling by converting the deterministic ODE sampler to an equivalent SDE sampler:
\begin{equation*}
    \begin{aligned}
\boldsymbol{x}_{t+\Delta t} &= \boldsymbol{x}_t + \left[ \boldsymbol{v}_\theta (\boldsymbol{x}_t, t) + \frac{\sigma_t^2}{2t} (\boldsymbol{x}_t + (1 - t)\boldsymbol{v}_\theta (\boldsymbol{x}_t, t)) \right] \Delta t \\
&\quad + \sigma_t \sqrt{\Delta t} \boldsymbol{\epsilon}
\end{aligned}
\end{equation*}
where $\sigma_t = a\sqrt{\frac{t}{1-t}}$, $a$ is a scalar hyper-parameter to guide the noise level, and $\boldsymbol{\epsilon} \sim \mathcal{N}(0, \boldsymbol{I})$ injects stochasticity. The KL divergence between $\pi_\theta$ and the reference policy $\pi_{\text{ref}}$ is a closed form:
\begin{equation*}
    \begin{aligned}
    D_{\text{KL}}(\pi_\theta \| \pi_{\text{ref}}) &= \frac{\left\| \bar{\boldsymbol{x}}_{t+\Delta t, \theta} - \bar{\boldsymbol{x}}_{t+\Delta t, \text{ref}} \right\|^2}{2\sigma_t^2 \Delta t} \\
    &= \frac{\Delta t}{2} \left( \frac{\sigma_t(1 - t)}{2t} + \frac{1}{\sigma_t} \right)^2  \\
    &\quad \times \left\| \boldsymbol{v}_\theta(\boldsymbol{x}_t, t) - \boldsymbol{v}_{\text{ref}}(\boldsymbol{x}_t, t) \right\|^2
    \end{aligned}    
\end{equation*}

\begin{algorithm}[tb]
\caption{Training Algorithm for CrystalGRPO-C}
\label{alg:crystalgrpo_c}
\small
\begin{algorithmic}[1]
\REQUIRE Pretrained policy $\pi_\theta$, reward $\mathcal{R}$,
group size $G$, batches per episode $B$
\ENSURE Optimized policy parameters $\theta$

\STATE Initialize and freeze
$\pi_{\mathrm{ref}}\leftarrow\pi_\theta$

\FOR{each training episode}
    \STATE Set behavior policy
    $\pi_{\mathrm{old}}\leftarrow\pi_\theta$

    \FOR{$b=1,\ldots,B$}
        \STATE Sample conditions $\mathcal{C}_b$ and initialize buffer
        $\mathcal{D}$

        \FOR{each $c\in\mathcal{C}_b$}
            \FOR{$g=1,\ldots,G$}
                \STATE Generate a joint coordinate--lattice trajectory
                $\tau_g$ using $\pi_{\mathrm{old}}$
                \STATE Store transition states, noises, and behavior means
                \STATE Compute $R_g=\mathcal{R}(\widehat X_g,c)$,
                validity $v_g$, and match flag $m_g$
            \ENDFOR

            \STATE Compute group-relative advantages $\{A_g\}_{g=1}^{G}$
            \STATE Set
            $\widetilde A_g=\max(A_g,0)$ if $v_g=1,m_g=0$;
            otherwise $\widetilde A_g=A_g$
            \STATE Add $\{(\tau_g,\widetilde A_g)\}_{g=1}^{G}$ to
            $\mathcal{D}$
        \ENDFOR

        \FOR{each policy optimization epoch}
            \STATE Compute policy ratios and full-trajectory
            reference regularization on $\mathcal{D}$
            \STATE Update $\theta$ by maximizing the clipped
            CrystalGRPO-C objective
        \ENDFOR
    \ENDFOR
\ENDFOR
\end{algorithmic}
\end{algorithm}

\begin{algorithm}[tb]
\caption{Training Algorithm for CrystalGRPO-Q}
\label{alg:crystalgrpo_q}
\small
\begin{algorithmic}[1]
\REQUIRE Pretrained policy $\pi_\theta$, reward $\mathcal{R}$,
group size $G$, batches per episode $B$
\ENSURE Optimized policy parameters $\theta$

\STATE Initialize and freeze
$\pi_{\mathrm{ref}}\leftarrow\pi_\theta$

\FOR{each training episode}
    \STATE Set behavior policy
    $\pi_{\mathrm{old}}\leftarrow\pi_\theta$

    \FOR{$b=1,\ldots,B$}
        \STATE Sample conditions $\mathcal{C}_b$ and initialize buffer
        $\mathcal{D}$

        \FOR{each $c\in\mathcal{C}_b$}
            \FOR{$g=1,\ldots,G$}
                \STATE Generate a joint coordinate--lattice trajectory
                $\tau_g$ using $\pi_{\mathrm{old}}$
                \STATE Store transition states, noises, and behavior means
                \STATE Compute terminal reward
                $R_g=\mathcal{R}(\widehat X_g,c)$
            \ENDFOR

            \STATE Compute group-relative advantages $\{A_g\}_{g=1}^{G}$
            \STATE Add $\{(\tau_g,A_g)\}_{g=1}^{G}$ to $\mathcal{D}$
        \ENDFOR

        \FOR{each policy optimization epoch}
            \STATE Compute policy ratios and early-stage
            reference regularization on $\mathcal{D}$
            \STATE Update $\theta$ by maximizing the clipped
            CrystalGRPO-Q objective
        \ENDFOR
    \ENDFOR
\ENDFOR
\end{algorithmic}
\end{algorithm}

\section{CrytalGRPO Training Algorithm}

\paragraph{Training procedure.}
The training procedures of CrystalGRPO-C and CrystalGRPO-Q are
summarized in Algorithms~\ref{alg:crystalgrpo_c} and
\ref{alg:crystalgrpo_q}, respectively. Both variants follow the
same on-policy GRPO pipeline, alternating between stochastic rollout
collection and policy optimization. A frozen reference policy
$\pi_{\mathrm{ref}}$ is initialized from the pretrained flow model
before training. At the beginning of each episode, the current policy
is copied to the behavior policy,
$\pi_{\mathrm{old}}\leftarrow\pi_{\theta}$, which is then used to
collect rollout trajectories.

For each conditioning instance $c$, we independently sample $G$
initial coordinate--lattice states and generate joint trajectories
$z_t^g=(F_t^g,L_t^g)$ using $\pi_{\mathrm{old}}$. At each generation
step, independent Gaussian noises are sampled for the coordinate and
lattice fields. The pre-wrapped coordinate proposal, lattice proposal,
sampled noises, and behavior-policy transition means are stored in the
rollout buffer. Storing these quantities allows each realized
transition to be re-evaluated under the updated policy without
resampling the trajectory.

At the terminal state, the generated crystal $\widehat X_g$ receives
a reward $R_g=\mathcal{R}(\widehat X_g,c)$. The general reward
formulation supports target-recovery guidance, energy guidance, or
their weighted combination, depending on the experimental
configuration. Rewards from candidates generated under the same
condition are normalized to obtain the group-relative advantage
$A_g$. The two CrystalGRPO variants differ in how this advantage is
used:
\begin{equation*}
\widetilde A_g^{\,\mathrm{Q}} = A_g,
\qquad
\widetilde A_g^{\,\mathrm{C}} =
\begin{cases}
\max(A_g,0), & v_g=1 \text{ and } m_g=0,\\
A_g, & \text{otherwise},
\end{cases}
\end{equation*}
where $v_g$ and $m_g$ denote the validity and reference-match flags,
respectively. CrystalGRPO-Q therefore retains the standard
group-relative update. CrystalGRPO-C prevents valid but currently
unmatched candidates from receiving negative updates solely because
their rewards fall below the group mean. Matched candidates retain
their original advantages and continue to compete according to
recovery quality, while invalid candidates remain suppressible.

During policy optimization, each stored transition is re-evaluated
under $\pi_{\theta}$ to compute the size-normalized surrogate policy
ratio $r_t(\theta)$. The two variants also use different temporal
schedules for reference-policy regularization:
\begin{equation*}
w_t^{\mathrm{Q}}=\mathbb{I}[t\geq\tau_{\mathrm{KL}}],
\qquad
w_t^{\mathrm{C}}=1.
\end{equation*}
CrystalGRPO-Q applies the reference constraint only during the early,
high-noise portion of generation, allowing more flexible
reward-guided refinement at later steps. CrystalGRPO-C applies the
constraint throughout the complete trajectory to more strongly
preserve the pretrained candidate distribution.

The unified objective for
$q\in\{\mathrm{Q},\mathrm{C}\}$ is
\begin{equation*}
\begin{aligned}
\mathcal{J}_{\mathrm{CG}}^{\,q}(\theta)
&=
\mathbb{E}_{c,g,t}
\left[
\ell_{g,t}^{\,q}(\theta)
\right],\\
\ell_{g,t}^{\,q}(\theta)
&=
\min\left(
r_t(\theta)\widetilde A_g^{\,q},
\bar r_t(\theta)\widetilde A_g^{\,q}
\right)
-
\beta w_t^{q}\widetilde D_{\mathrm{KL},t},\\
\bar r_t(\theta)
&=
\operatorname{clip}
\left(
r_t(\theta),1-\epsilon,1+\epsilon
\right),
\end{aligned}
\label{eq:crystalgrpo_objective}
\end{equation*}
where $\epsilon$ is the PPO clipping coefficient, $\beta$ controls
reference-policy regularization, and
$\widetilde D_{\mathrm{KL},t}$ is the size-normalized divergence from
the frozen reference policy.

\begin{figure*}[t]
    \centering
    \includegraphics[width=\textwidth]{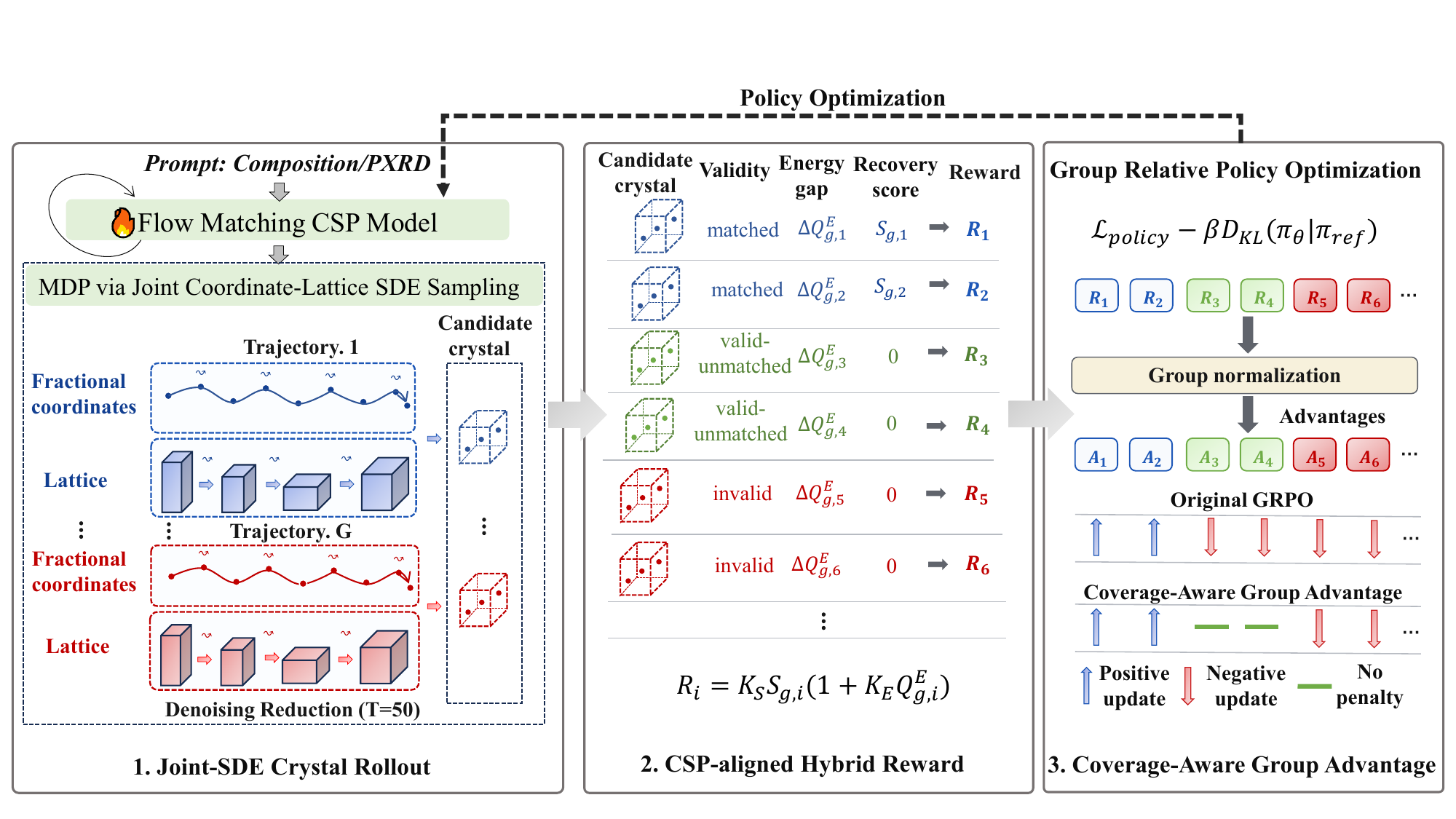}
    \caption{
    Overview of the CrystalGRPO framework.
    CrystalGRPO performs joint stochastic rollouts over fractional coordinates and lattice variables, evaluates the generated crystals using a CSP-aligned hybrid reward. Based on the hybrid reward, it computes coverage-aware group advantages for GRPO policy optimization.
    }
    \label{fig:method_overview}
\end{figure*}


\subsection{Comparison between CrystalGRPO-Q and CrystalGRPO-C}

CrystalGRPO-Q and CrystalGRPO-C share the same joint
coordinate--lattice SDE policy and use the same reward configuration
within each experiment. Their differences lie in the temporal schedule
of reference regularization and the treatment of valid but unmatched
candidates, as summarized in Table~\ref{tab:q_c_comparison}.

\begin{table}[t]
\centering
\caption{Comparison between CrystalGRPO-Q and CrystalGRPO-C.}
\label{tab:q_c_comparison}
\resizebox{\columnwidth}{!}{
\begin{tabular}{lcc}
\toprule
Component & CrystalGRPO-Q & CrystalGRPO-C \\
\midrule
Joint coordinate--lattice SDE & Yes & Yes \\
Reward configuration & Shared within each experiment
                     & Shared within each experiment \\
Frozen pretrained policy & Yes & Yes \\
Reference KL schedule
& Early high-noise steps & Full trajectory \\
Group advantage
& Standard $A_g$ & Coverage-aware $\widetilde A_g$ \\
Valid unmatched candidate
& May receive $A_g<0$
& Set $\widetilde A_g=\max(A_g,0)$ \\
Reward-driven concentration
& Less constrained & More constrained \\
Intended operating point
& Single-draw recovery & Finite-budget recovery \\

\bottomrule
\end{tabular}}
\end{table}

CrystalGRPO-Q applies reference regularization only during the early,
high-noise portion of the trajectory and retains the standard
group-relative advantage. This allows greater flexibility during later
structural refinement, but may concentrate probability mass around a
smaller number of high-reward modes. CrystalGRPO-C instead applies
reference regularization throughout the trajectory and removes negative
advantages for valid unmatched candidates. It therefore limits policy
drift and avoids suppressing potentially recoverable structural basins
solely because their current rewards fall below the group mean.

\section{Detailed Reward Design and Experimental Configurations}
\label{sec:reward_details}

\subsection{CSP-aligned Hybrid Reward}

For completeness and reproducibility, we provide the exact reward
implementation used to obtain the reported results. In the
target-aligned reward configuration, the StructureMatcher score
provides the structural anchor, while the bounded MACE energy term
refines the ranking only among candidates that pass the recovery
matcher.

\paragraph{Per-atom energy gap.}

Let $\mathcal{E}_{\mathrm{MACE}}(X)$ denote the total energy in eV
predicted by MACE-MP-0 \cite{batatia2025foundation}. We first convert it
to an energy per atom,
\begin{equation*}
\varepsilon(X)
=
\frac{\mathcal{E}_{\mathrm{MACE}}(X)}{N_X},
\label{eq:energy_per_atom}
\end{equation*}
where $N_X$ is the number of atoms in the structure. All energy
quantities below are therefore expressed in
$\mathrm{eV\,atom^{-1}}$.

For generated structures, the predicted energy is numerically bounded
as
\begin{equation*}
\bar{\varepsilon}(\widehat X_g)
=
\operatorname{clip}
\left(
\varepsilon(\widehat X_g),-30,3
\right).
\label{eq:pred_energy_clip}
\end{equation*}
Structures that fail geometric-validity checks or MACE evaluation are
assigned the upper value of $3~\mathrm{eV\,atom^{-1}}$. Ground-truth
energies are precomputed using the same MACE model and stored as
per-atom values before training.
For a finite cached ground-truth energy, the per-atom energy gap is
\begin{equation*}
\Delta\varepsilon_g
=
\bar{\varepsilon}(\widehat X_g)
-
\varepsilon(X_{\mathrm{GT}}).
\label{eq:energy_gap}
\end{equation*}
For energy-only experiments, we use
\begin{equation}
R_g^{E}
=
-\widetilde{\Delta\varepsilon}_g,
\end{equation}
where
$\widetilde{\Delta\varepsilon}_g
=\operatorname{clip}(\Delta\varepsilon_g,-15,3)$.

For the setting for hybrid reward, we then convert the energy gap into a dimensionless bounded quality
score:
\begin{equation*}
Q_g^{E}
=
\operatorname{clip}
\left(
-\frac{\widetilde{\Delta\varepsilon}_g}{E_0},
-1,1
\right),
\qquad
E_0=0.5~\mathrm{eV\,atom^{-1}}.
\label{eq:bounded_energy_quality}
\end{equation*}
Thus, a prediction at least $0.5~\mathrm{eV\,atom^{-1}}$ below the
ground-truth energy receives $Q_g^{E}=1$, an equal-energy prediction
receives $Q_g^{E}=0$, and a prediction at least
$0.5~\mathrm{eV\,atom^{-1}}$ above the ground truth receives
$Q_g^{E}=-1$. This fixed clipping prevents spuriously low MACE energies
on out-of-distribution geometries from dominating the policy update.

For the rare samples whose ground-truth MACE energy cannot be
precomputed, the implementation falls back to the absolute predicted
per-atom energy. In MP-20 this affects 19 of 27,136 training structures
($0.07\%$).

\paragraph{StructureMatcher recovery sco.}

The reported structural metric and the strict reward gate use
\texttt{StructureMatcher} \cite{ong2013python} with  \texttt{stol}$=0.5$, \texttt{ltol}$=0.3$, and
\texttt{angle\_tol}$=10^\circ$, 
Both the generated and ground-truth structures must pass the same
composition and structural-validity checks used during evaluation.

Let $m_g\in\{0,1\}$ indicate whether this matcher succeeds. When
$m_g=1$, we use \texttt{StructureMatcher} to get the normalized RMS displacement as $d_g$. Furthermore, The recovery score can be defined as:
\begin{equation*}
S_g
=
\begin{cases}
\exp\left(-d_g/\tau\right), & m_g=1,\\
0, & m_g=0,
\end{cases}
\qquad
\tau=0.5.
\label{eq:recovery_score_operational}
\end{equation*}
Here $\tau$ controls the sensitivity to the normalized structural
displacement and is distinct from the matcher parameter
$\texttt{stol}$, and a structure that fails the reported evaluation matcher receives $S_g=0$.

\paragraph{Final hybrid reward.}

The final hybrid reward combines the energy signal with the recovery score as:
\begin{equation*}
R_g
=
K_S S_g
\left(
1+K_E Q_g^{E}
\right),
\label{eq:gated_hybrid_reward}
\end{equation*}
Because we set $S_g=0$ for an unmatched candidate, the energy term acts
only as a bounded refinement after target recovery. 

\subsection{Common RL Hyperparameters}
\label{sec:common_rl_hyperparameters}

Unless otherwise specified, all CrystalGRPO experiments use the common
hyperparameters summarized in Table~\ref{tab:common_rl_hparams}. We optimize
the policy using AdamW with a constant learning rate of $10^{-5}$ and no
weight decay. Each training episode contains three mini-batches of 256 target
crystals. For every target crystal, we independently sample $G=12$ rollouts,
corresponding to $3\times256\times12=9{,}216$ generated structures per
episode. Both training rollouts and deterministic evaluation use 50 numerical
integration steps. All models were implemented in PyTorch and trained on NVIDIA H100 and H200 GPUs.

The PPO clipping parameter $\epsilon=0.2$ restricts each local importance
ratio to the interval $[0.8,1.2]$, and old policy is refreshed at the
beginning of every episode to remain fixed during all updates in that
episode. Rewards are normalized independently over the $G$ rollouts belonging to the same target crystal.

For stochastic rollouts, both the coordinate and the lattice diffusion scale follows the schedule below:
\begin{equation}
    \sigma_(t)
    =
    \min\left(
    0.2\sqrt{\frac{t}{1-t}},\,5
    \right).
\end{equation}
The reference regularizer is computed from the normalized squared
difference between the current and frozen-reference transition
means, corresponding to the Gaussian divergence defined in the main
paper. Its coefficient and active time interval depend on the
optimization configuration.

For the hybrid reward, all experiments use MACE-MP-0 ``medium'' to predict
per-atom energies. The structural reward weight is $K_S=1$, the gated
energy coefficient is $K_E=0.05$, the geometric decay scale is
$\tau=0.5$, and the energy normalization scale is $E_0=0.5$ eV/atom.
Consequently, the energy term changes the reward of an accepted structure by
at most $\pm5\%$. Unmatched structures retain exactly zero reward. Predicted
per-atom energies are bounded to $[-30,3]$ eV/atom, while the energy
difference $E(\widehat X_g)-E(X_{\mathrm{GT}})$ is bounded to
$[-15,3]$ eV/atom before constructing the normalized energy-quality score.
Failed or invalid MACE evaluations are assigned the upper-bound value of
$3$ eV/atom.

\begin{table}[t]
\centering
\caption{Common RL hyperparameters used in CrystalGRPO.}
\label{tab:common_rl_hparams}
\small
\begin{tabular}{ll}
\toprule
\textbf{Hyperparameter} & \textbf{Value} \\
\midrule
Learning rate $\eta$ & $1\times10^{-5}$ \\
Maximum training episodes & $3{,}000$ \\
Mini-batch size $M$ & $256$ crystals \\
batches per episode $B$ & $3$ \\
Rollouts per crystal $G$ & $12$ \\
Training integration steps & $50$ \\
Noise level & $0.2$ \\
PPO clipping parameter $\epsilon$ & $0.2$ \\
Advantage clipping range & $[-5,5]$ \\
Ratio rollback threshold & $1.5$ \\
Energy model & MACE-MP-0 medium \\
$K_S$ & $1.0$ \\
$K_E$ & $0.05$ \\
Geometric decay scale $\tau$ & $0.5$ \\
Energy normalization scale $E_0$ & $0.5$ eV/atom \\
\bottomrule
\end{tabular}
\end{table}

\begin{table}[t]
\centering
\caption{Dataset-, backbone-, and optimization-specific configurations.}
\label{tab:specific_rl_hparams}
\small
\begin{tabular}{p{0.38\linewidth}p{0.54\linewidth}}
\toprule
\textbf{Hyperparameter} & \textbf{Value} \\
\midrule
PXRDGen backbone initialization
& Dataset-specific pretrained PXRDGen checkpoint \\

OMatG backbone initialization
& Dataset-specific ported OMatG checkpoint \\

Fractional-coordinate prior
& $\mathcal{U}[0,1)$ for both backbones \\

PXRDGen lattice prior
& Isotropic Gaussian base distribution \\

OMatG lattice prior
& Dataset-dependent informed lattice prior \\
Advantage mode for CrystalGRPO-Q
& Normal advantage \\

KL active interval for CrystalGRPO-Q
& $t\geq0.6$ \\

Advantage mode for CrystalGRPO-C
& Coverage-preserving advantage \\

KL active interval for CrystalGRPO-C
& $t\geq0.0$ \\

\bottomrule
\end{tabular}
\end{table}

\subsection{Dataset- and Backbone-Specific Configurations}
\label{sec:specific_rl_hyperparameters}

Table~\ref{tab:specific_rl_hparams} lists the settings that are not shared by
all experiments. The PXRDGen-initialized models use the corresponding
pretrained MP-20 or MPTS-52 checkpoint. Their initial fractional coordinates
are sampled uniformly in the unit cell, while the initial lattice is sampled
from the Gaussian base distribution used during flow-matching pretraining.

For the OMatG-backed \cite{omatg} experiments, we initialize the policy from the
released dataset-specific Linear-ODE checkpoint. Specifically, we extract
the pretrained CSPNet velocity-field parameters and load them into an
architecture-equivalent adapter that exposes the same CrystalGRPO policy
interface. The adapter preserves OMatG's fractional-coordinate and
$3\times3$ lattice representations, sinusoidal time embedding under the
mapping $s=1-\tau$, center-of-mass correction, dataset-specific informed
lattice prior, and pretrained coordinate and lattice velocity fields.

During RL training, CrystalGRPO converts the pretrained ODE velocity field
into Gaussian SDE transitions using its fixed diffusion schedule. GRPO
advantages and per-step PPO ratios are then used to directly fine-tune the
CSPNet parameters, while a frozen copy of the pretrained model provides the
KL reference. Thus, no separate velocity-correction network is introduced;
policy optimization modifies the underlying velocity field itself. Final
evaluation is performed by deterministic ODE integration of the fine-tuned
velocity field.

\subsection{Reward configurations used in the reported experiment}
Table ~\ref{tab:reward_setting} lists the reward configurations used in the reported experiment, where the Composition-conditioned Q/C results are based on the hybrid reward we defined, while the energy-only reward ablation and PXRD-conditioned studies are based on energy-only reward.
\begin{table}[t]
\centering
\caption{Reward configurations used in the reported experiments.}
\label{tab:reward_setting}
\begin{tabular}{lc}
\toprule
Experiment & Reward configuration \\
\midrule
Composition-conditioned Q/C results
& Hybrid reward \\
Energy-only reward ablation
& Energy-only reward \\
PXRD-conditioned analysis
& Energy-only reward \\
\bottomrule
\end{tabular}
\end{table}

\end{document}